\documentclass[fleqn,10pt]{wlscirep}
\usepackage[utf8]{inputenc}
\usepackage[T1]{fontenc}

\usepackage{amsmath,amssymb,amsfonts}
\usepackage{svg}
\usepackage{dsfont}
\usepackage{graphicx}
\usepackage{textcomp}
\usepackage{xcolor}
\usepackage{hyperref}       
\usepackage{url}            
\usepackage{booktabs}       
\usepackage{nicefrac}       
\usepackage{microtype}      
\usepackage{subfigure}
\usepackage{epsfig}
\usepackage{epstopdf} 
\usepackage{tabularx}
 \usepackage{algpseudocode}
\usepackage{algorithm}
\usepackage{multirow}
\usepackage{subfiles}

\usepackage{xr}

\newcommand{\beginsupplement}{%
        \setcounter{table}{0}
        \renewcommand{\thetable}{S\arabic{table}}%
        \setcounter{figure}{0}
        \renewcommand{\thefigure}{S\arabic{figure}}%
        \setcounter{section}{0}
         \renewcommand{\thesection}{S\arabic{section}}
\renewcommand{\theequation}{S\arabic{equation}}
}

\makeatletter
\newcommand*{\addFileDependency}[1]{
  \typeout{(#1)}
  \@addtofilelist{#1}
  \IfFileExists{#1}{}{\typeout{No file #1.}}
}
\makeatother

\title{Multimodal sensor fusion in the latent representation space}

\author[1,*]{Robert J. Piechocki}
\author[1]{Xiaoyang Wang}
\author[1]{Mohammud J. Bocus}

\affil[1]{School of Computer Science, Electrical and Electronic Engineering, and Engineering Maths\\
  University of Bristol, Bristol, BS8 1UB, UK.}

\affil[*]{r.j.piechocki@bristol.ac.uk}

\begin{abstract}
A new method for multimodal sensor fusion is introduced. The technique relies on a two-stage process. In the first stage, a multimodal generative model is constructed from unlabelled training data. In the second stage, the generative model serves as a reconstruction prior and the search manifold for the sensor fusion tasks. The method also handles cases where observations are accessed only via subsampling i.e. compressed sensing. We demonstrate the effectiveness and excellent performance on a range of multimodal fusion experiments such as multisensory classification, denoising, and recovery from subsampled observations.
\end{abstract}
\begin{document}

\flushbottom
\maketitle
%
%
\thispagestyle{empty}


\section*{Introduction}
\emph{Controlled hallucination}\cite{ASeth} 
is an evocative term referring to the Bayesian brain hypothesis \cite{BayesBrain}. It posits that perception is not merely a function of sensory information processing capturing the world as is. Instead, the brain is a predictive machine - it attempts to infer the causes of sensory inputs. To achieve this, the brain builds and continually refines its world model. The world model serves as a prior and when combined with the sensory signals will produce the best guess for its causes. Hallucination (uncontrolled) occurs when the sensory inputs cannot be reconciled with, or contradict the prior world model. This might occur in our model, and when it does, it manifests itself at the fusion stage with the stochastic gradient descent procedure getting trapped in a local minimum. The method presented in this paper is somewhat inspired by the Bayesian brain hypothesis, but it also builds upon multimodal generative modelling and deep compressed sensing.

Multimodal data fusion attracts academic and industrial interests alike \cite{survey_fusion} and plays a vital role in several applications. Automated driving is arguably the most challenging industrial domain \cite{FusionDriving}. Automated vehicles use a plethora of sensors: Lidar, mmWave radar, video and ultrasonic, and attempt to perform some form of sensor fusion for environmental perception and precise localization. A high-quality of final fusion estimate is a prerequisite for safe driving. 
Amongst other application areas, a notable mention deserves eHealth and Ambient Assisted Living (AAL). These new paradigms are contingent on gathering information from various sensors around the home to monitor and track the movement signatures of people. The aim is to build long-term behavioral sensing machine which also affords privacy. Such platforms rely on an array of environmental and wearable sensors, with sensor fusion being one of the key challenges. 

\begin{figure}
    \centering
    \includegraphics[width=\linewidth]{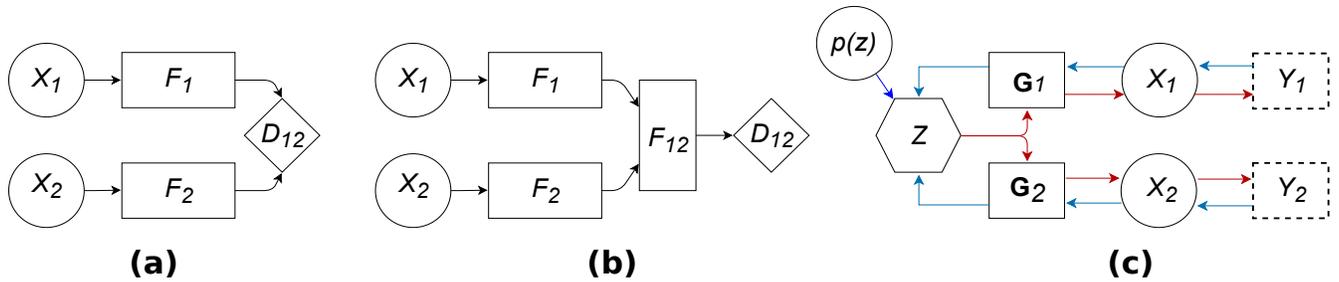} 
    \caption{Multimodal Sensor Fusion: (\textbf{a}) Decision fusion, (\textbf{b}) Feature fusion, (\textbf{c}) Our technique: fusion in the latent representation with optional compressed sensing measurements; $F$ features, $p(z)$ prior model, $\bf{G}$ generators, $X$ complete data, $Y$ subsampled data. For clarity $M=2$ modalities are shown, the concept generalises to any $M$.}
    \label{fig:SensorFusion}
\end{figure}

In this contribution, we focus on a one-time snapshot problem (i.e. we are not building temporal structures). However, we try to explore the problem of multimodal sensor fusion from a new perspective, essentially, from a Bayesian viewpoint.The concept is depicted in Fig.~\ref{fig:SensorFusion}, alongside two main groups of approaches to sensor fusion. Traditionally, sensor fusion for classification tasks has been performed at the decision level as in Fig.~\ref{fig:SensorFusion}(a). Assuming that conditional independence holds, a pointwise product of final pmf (probability mass function) across all modalities is taken. Feature fusion, as depicted in Fig.~\ref{fig:SensorFusion}(b), has become very popular with the advent of deep neural networks \cite{survey_fusion}, and can produce very good results. Fig.~\ref{fig:SensorFusion}(c) shows our technique during the fusion stage (Stage 2). Blue arrows indicate the direction of backpropagation gradient flow during fusion.          

\textbf{Contributions:}
\begin{itemize}
\item 
A novel method for multimodal sensor fusion is presented. The method attempts to find the best estimate (\emph{maximum a posteriori}) for the causes of observed data. The estimate is then used to perform specific downstream fusion tasks.
\item
The method can fuse the modalities under lossy data conditions i.e. when the data is subsampled, lost and/or noisy. Such phenomena occur in real-world situations such as the transmission of information wirelessly, or intentional subsampling to expedite the measurement (rapid MRI imaging and radar) etc. 
\item
It can leverage between modalities. A strong modality can be used to aid the recovery of another modality that is lossy or less informative (weak modality). This is referred to as asymmetric Compressed Sensing.  
\end{itemize}

\section*{Related Work}
In this section, we review the state-of-the-art in three areas directly relevant to our contribution: multimodal generative modeling, sensor fusion, and compressed sensing. One of the main aims of Multimodal Variational Autoencoders (MVAEs) is to learn shared representation across different data types in a fully self-supervised manner, thus avoiding the need to label a huge amount of data, which is time-consuming and expensive \cite{mopoe}. It is indeed a challenge to infer the low-dimensional joint representation from multiple modalities, which can ultimately be used in downstream tasks such as self-supervised clustering or classification. This is because the modalities may vastly differ in characteristics, including dimensionality, data distribution, and sparsity \cite{moeproteindata}.
Recently, several methods have been proposed to combine multimodal data using generative models such as Variational Autoencoders (VAEs) \cite{mvae, mvae_moe, mopoe, jmvae, dmvae, cadavae}. These methods aim to learn a joint distribution in the latent space via inference networks and try to reconstruct modality-specific data, even when one modality is missing. In these works, a modality can refer to natural images, text, captions, labels or visual and non-visual attributes of a person. JMVAE (Joint Multimodal Variational Autoencoder) \cite{ jmvae} makes use of a joint inference network to learn the interaction between two modalities and they address the issue of missing modality by training an individual (unimodal) inference network for each modality as well as a bimodal inference network to learn the joint posterior, based on the product-of-experts (PoE). They consequently minimize the distance between unimodal and multimodal latent distribution. On the other hand, MVAE \cite{mvae}, which is also based on PoE, considers only a partial combination of observed modalities, thereby reducing the number of parameters and improving the computational efficiency. Reference \cite{mvae_moe} uses the Mixture-of-Experts (MoE) approach to learn the shared representation across multiple modalities. The latter two models essentially differ in their choices of joint posterior approximation functions. 
MoPoE (Mixture-of-Products-of-Experts)-VAE \cite{mopoe} aims to combine the advantages of both approaches, MoE and PoE, without incurring significant trade-offs. DMVAE (Disentangled Multimodal VAE) \cite{dmvae} uses a disentangled VAE approach to split up the private and shared (using PoE) latent spaces of multiple modalities, where the latent factor may be of both continuous and discrete nature. CADA (Cross- and Distribution Aligned)-VAE \cite{cadavae} uses a cross-modal embedding framework to learn a latent representation from image features and classes (labels) using aligned VAEs optimized with cross- and distribution- alignment objectives.

In terms of multimodal/sensor fusion for human activity sensing using Radio-Frequency (RF), inertial and/or vision sensors, most works have considered either decision-level fusion or feature-level fusion. For instance, the work in \cite{wivi} performs multimodal fusion at the decision level to combine the benefits of WiFi and vision-based sensors using a hybrid deep neural network (DNN) model to achieve good activity recognition accuracy for 3 activities. The model essentially consists of a WiFi sensing module (dedicated Convolutional Neural Network (CNN) architecture) and a vision sensing module (based on the Convolutional 3D model) for processing WiFi and video frames for unimodal inference, followed by a multimodal fusion module. Multimodal fusion is performed at the decision level (after both WiFi and vision modules have made a classification) because this framework is stated to be more flexible and robust to unimodal failure compared to feature level fusion. Reference~\cite{gimme} presents a method for activity recognition, which leverages four sensor modalities, namely, skeleton sequences, inertial and motion capture measurements and WiFi fingerprints.
The fusion of signals is formulated as a matrix concatenation. The individual signals of different sensor modalities are transformed and represented as an image. The resulting images are then fed to a two-dimensional CNN (EfficientNet B2) for classification. 
The authors of \cite{wiwehar} proposed a multimodal HAR system that leverages WiFi and wearable sensor modalities to jointly infer human activities. They collect Channel Sate Information (CSI) data from a standard WiFi Network Interface Card (NIC), alongside the user's local body movements via a wearable Inertial Measurement Unit (IMU) consisting of an accelerometer, gyroscope, and magnetometer sensors. They compute the time-variant Mean Doppler Shift (MDS) from the processed CSI data and magnitude from the inertial data for each sensor of the IMU. Then, various time and frequency domain features are separately extracted from the magnitude data and the MDS. The authors apply a feature-level fusion method which sequentially concatenates feature vectors that belong to the same activity sample. Finally supervised machine learning techniques are used to classify four activities, such as walking, falling, sitting, and picking up an object from the floor.

Compared to the aforementioned works \cite{ wivi, gimme, wiwehar} which consider supervised models with feature-level fusion or decision-level fusion, our technique, in contrast, performs multimodal sensor fusion in the latent
representation space leveraging a self-supervised generative model. Our method is different from current multimodal generative models such as those proposed in \cite{mvae, mvae_moe, mopoe, jmvae} in the sense that it can handle cases where observations are accessed only via subsampling (i.e. compressed sensing with significant loss of data and no data imputation). And crucially our technique attempts to directly compute the MAP (\emph{maximum a posteriori}) estimate.

The presented method is related to and builds upon Deep Compressed Sensing (DCS) techniques\cite{csgm, DeepCompSens}. DCS, in turn, is inspired by Compressed Sensing (CS) \cite{Candes-CS, Donoho-CS}. In CS, we attempt to solve what appears to be an underdetermined linear system, yet the solution is possible with the additional prior sparsity constraint on the signal: $\min L0$. Since $L0$ is non-convex, $L1$ is used instead to provide a convex relaxation, which also promotes sparsity and allows for computationally efficient solvers. DCS, in essence, replaces the $L0$ prior with a low dimensional manifold, which is learnable from the data using generative models. Concurrently to DCS, Deep Image Prior \cite{DeepImagePrior_2018_CVPR} was proposed. It used un-trained CNNs to solve a range of inverse problems in computer vision (image inpainting, super-resolution, denoising).

\section*{Methods}
Assume the data generative process so that latent and common cause $Z$ gives rise to $X_m$, which in turn produces observed $Y_m$, i.e. $Z\rightarrow X_m \rightarrow Y_m $ forms a Markov chain.  Here, $X_m$ is the full data pertaining to $m^{th}$ modality, $m \in \{1,\dots,M\}$. Crucially, the modalities collect data simultaneously ``observing'' the same scene. As an example, in this work, we consider the
different views (obtained via multiple receivers) from the opportunistic CSI WiFi radar as different modalities.
The variable $Z$ encodes the semantic content of the scene and is typically of central interest. Furthermore, $X_m$ is not accessed directly, but is observed via a \emph{subsampled} $Y_m$. This is a compressed sensing setup: $Y_m=\chi_{m}(X_m)$: $\chi_{m}$ is a deterministic and known (typically many-to-one) function. The only condition we impose on $\chi_{m}$ is to be Lipschitz continuous.  With the above, the conditional independence between modalities holds (conditioned on $Z$). Therefore, the joint density factors as:   

\begin{equation}
p\left(z,x_{1:M},y_{1:M}\right) = p\left( z\right) \prod_{m=1}^M{ p(y_{m}|x_{m}) p(x_{m}|z) }.
\label{eq:main_model}
\end{equation}

The main task in this context is to produce the best guess for latent $Z$, and possibly, to recover the full signal(s) $X_m$, given subsampled data $Y_{1:M}$. We approach the problem in two stages. First we build a joint model which approximates equation \eqref{eq:main_model}, and will be instantiated as a Multimodal Variatational Autoencoer (M-VAE). More specifically, the M-VAE will provide an approximation to $p_{\phi_{1:M},\psi_{1:M}}(z,x_{1:M})$, parameterized by deep neural networks $\{\phi_{1},\dots, \phi_{M}\}$, $\{\psi_{1},\dots, \psi_{M}\}$, referred to as \emph{encoders} and \emph{decoders}, respectively. The trained M-VAE will then be appended with $p_{\chi_{m}}(y_{m}|x_{m})$ for each modality $m$: $\{\chi_{1},\dots, \chi_{M}\}$ referred to as \emph{samplers}. 
In the second stage, we use the trained M-VAE and $\chi_{1:M}$ to facilitate the fusion and reconstruction tasks. Specifically, our sensor fusion problem amounts to finding the maximum a posteriori (MAP) $\hat{z}_{MAP}$ estimate of the latent cause for a given ($i^{th}$) data point $Y_{1:M}=y^{(i)}_{1:M}$:

\begin{equation}
\hat{z}_{MAP} = \arg\max_{z}p\left(z| Y_{1:M}=y^{(i)}_{1:M} \right),
\label{eq:map_estimate}
\end{equation}
where, 
\begin{equation}
p\left(z| Y_{1:M}=y^{(i)}_{1:M} \right) \propto p\left( z\right) \prod_{m=1}^M{\int_{X_{m}}  p( Y_{m}=y^{(i)}_{m} |x_{m}) p(x_{m}|z) \,dx_{m}}.
\label{eq:map_mainpdf}
\end{equation}

The above MAP estimation problem is hard, and we will resort to approximations detailed in the sections below.

\subsection*{Multimodal VAE}
The first task is to build a model of equation  \eqref{eq:main_model}. As aforementioned, this will be accomplished in two steps. Firstly, during the training stage we assume access to full data $X_{1:M}$, therefore training an approximation to $p_{\phi_{1:M},\psi_{1:M}}(z,x_{1:M})$ is a feasible task.     
The marginal data log-likelihood for the multimodal case is:

\begin{align}
    \ \log p(x_{1:M}) &= D_{KL}(q(z|x_{1:M}||p(z|x_{1:M})) \\ &+ 
    \left[ \sum_{X_{m}} { \mathbb{E}_{z \sim q(z|x_{1:M})} \log p(x_{m}|z)}  - \mathbb{E}_{z \sim q(z|x_{1:M})}  \log \frac{q(z|x_{1:M})}{p(z)} \right] ,
\label{eq:ELBO}
\end{align}
where $D_{KL}$ is the Kullback–Leibler (KL) divergence.
The first summand in equation \eqref{eq:ELBO}, i.e. the sum over modalities follows directly from the conditional independence. And since KL is non-negative, equation (\ref{eq:ELBO}) represents the lower bound (also known as Evidence Lower Bound - ELBO) on the log probability of the data (and its negative is used as the loss for the M-VAE). There exist a body of work on M-VAEs, the interested reader is referred to \cite{mvae, mvae_moe, mopoe, jmvae} for details and derivation.
The key challenge in training M-VAEs is the construction of variational posterior $q(z|x_{1:M})$.
We dedicate a section in the Supplementary Information document \ref{App:PoE} to the discussion on choices and implications for the approximation of variational posterior. Briefly, we consider two main cases: a missing data case – i.e. where particular modality data might be missing ($X_m = x_m^{(i)} = \emptyset $); and the full data case. The latter is straightforward and is tackled by enforcing a particular structure of the encoders. For the former case variational Product-of-Experts (PoE) is used:



\begin{equation}
q_{\Phi}(z|x_{1:M}) = p(z) \prod_{m=1}^M{q_{\phi_m}(z|x_{m})}   .
\label{eq:PoE}
\end{equation}
Should the data be missing for any particular modality, $q_{\phi_m}(z|x_m) = 1$ is assumed. Derivation of equation~\eqref{eq:PoE} can be found in the Supplementary Information document \ref{App:PoE}.

\subsection*{Fusion on the M-VAE prior}

Recall the sensor fusion problem as stated in equation \eqref{eq:map_estimate}. The $p(z)$ is forced to be isotropic Gaussian by M-VAE, and the remaining densities are assumed to be Gaussian. Furthermore, we assume that $p(x_{m}|z)= \delta(x_{m}-\psi_{m}(z))$. Therefore equation \eqref{eq:map_estimate} becomes:     
\begin{equation}
\hat{z}_{MAP} = \arg\max_{z} p\left(z| Y_{1:M}=y^{(i)}_{1:M} \right) \propto \exp{(- \|z\|^2 )} \prod_{m=1}^M{\exp{(-\frac{1}{2\sigma_{m}^{2}} \|y_{m}^{(i)}- \chi_{m}(\psi_{m}(z))\|^2 )}}.
\label{eq:map_pfd}
\end{equation}
Hence, the objective to minimize becomes: 
\begin{equation}
\mathcal{L}(z) =    \lambda_0{\|z\|^2} + \sum_{m=1}^M{\lambda_m \|y_{m}^{(i)}- \chi_{m}(\psi_{m}(z))\|^2 }.
\label{eq:map_loss}
\end{equation}
Recall that the output of the first stage is $p(z)$ and the decoders $\prod_{m} {p_{\psi_{n}}(x|z)}$ are parametrized by $\{\psi_{1:M}\}$, $\{\lambda_{0:M}\}$ are constants.   
The MAP estimation procedure consists of backpropagating through the sampler $\chi_m$ and decoder $\psi_{m}$ using Stochastic Gradient Descent (SGD). 
In this step $\{\psi_{1:M}\}$ are non-learnable, i.e. jointly with $\chi_m$ are some non-linear known (but differentiable) functions.  
\begin{equation}
 z\leftarrow z - \eta_0\nabla_{z} ({{\|z\|^2}}) - \sum_{m=1}^M{\eta_m \nabla_{z}( \|y_{m}^{(i)}- \chi_{m}(\psi_{m}(z))\|^2) }.
\end{equation}
The iterative fusion procedure is initialized by taking a sample from the prior $z^{0} \sim p(z)$, $\{\eta_{0:M}\}$ are learning rates. One or several SGD steps are taken for each modality in turn. The procedure terminates with convergence - see Algorithm \ref{alg:cap}.  
In general, the optimization problem as set out in equation \eqref{eq:map_loss} is non-convex. Therefore, there are no guarantees of convergence to the optimal point ($\hat{z}_{MAP} $). We deploy several strategies to minimize the risk of getting stuck in a local minimum. We consider multiple initialization points (a number of points sampled from the prior with Stage 2 replicated for all points). In some cases it might be possible to sample from: $z^{0} \sim p\left(z \right)\prod p\left(z\left | X = \check{x}_m^{(j)} \right.\right)$. Depending on modality, this might be possible with data imputation ($\check{x}_{m}$ are imputed data). The final stage will depend on a particular task (multisensory classification/reconstruction), but in all cases it will take $\hat{z}_{MAP}$ as an input. In our experiments, we observe that the success of Stage 2 is crucially dependent on the quality of M-VAE.

\begin{algorithm}
    \caption{Multimodal Sensor Fusion in the Latent Representation Space (SFLR)}
    \label{alg:cap}
    \begin{algorithmic}[1]
\State {Training data: $\mathcal{D_{T}} \equiv \{ X_{1:M}^{(1:I)} \}$, Test data $\mathcal{D_{P}} \equiv \{ X_{1:M}^{(1:J)} \}$, Samplers $\{\chi_{1:M}\}$ }

\State {\textbf{Stage 1}: Train M-VAE using $\mathcal{D_{T}}$}

\State {Output:  $p(z)$, Encoders $\{\phi_{1:M} \}$, Decoders $\{ \psi_{1:M}\}$ }

\State {\textbf{Stage 2}: Fusion}

\State {$y_{1:M}^{(i)} \sim \mathcal{D_{P}} $} 

\State {Sample the initial point $z^{0} \sim p(z)$}

\While{not converged}
    \State $z\leftarrow z - \eta_0\nabla_{z} ({{\|z\|^2}}) - {\eta_1 \nabla_{z}( \|y_{1}^{(i)}-\chi_{1}(\psi_{1}(z))\|^2) }$
    
    \State $z\leftarrow z - \eta_0\nabla_{z} ({{\|z\|^2}}) - {\eta_2 \nabla_{z}( \|y_{2}^{(i)}-\chi_{2}(\psi_{2}(z))\|^2) }$
    
    \State $\vdots$
    
    \State $z\leftarrow z - \eta_0\nabla_{z} ({{\|z\|^2}}) - {\eta_M \nabla_{z}( \|y_{M}^{(i)}-\chi_{M}(\psi_{M}(z))\|^2) }$

\EndWhile
\State $\hat{z}_{MAP} \leftarrow z$
\State Downstream tasks: $\hat{x}_{m} = \psi_{m}(\hat{z}_{MAP})$, classification tasks $K$-NN$(\hat{z}_{MAP})$     
    \end{algorithmic}
\end{algorithm}

\section*{Experiments}

In this work, we investigate the performance of the proposed method on two datasets for multimodal sensor fusion and recovery tasks: i) a synthetic ``toy protein'' dataset and ii) a passive WiFi radar dataset intended for Human Activity Recognition (HAR).  

\subsection*{Synthetic toy protein dataset}
\label{sec:protein_data}
\begin{figure}[htb]
    \centering
    \includegraphics[width=\linewidth]{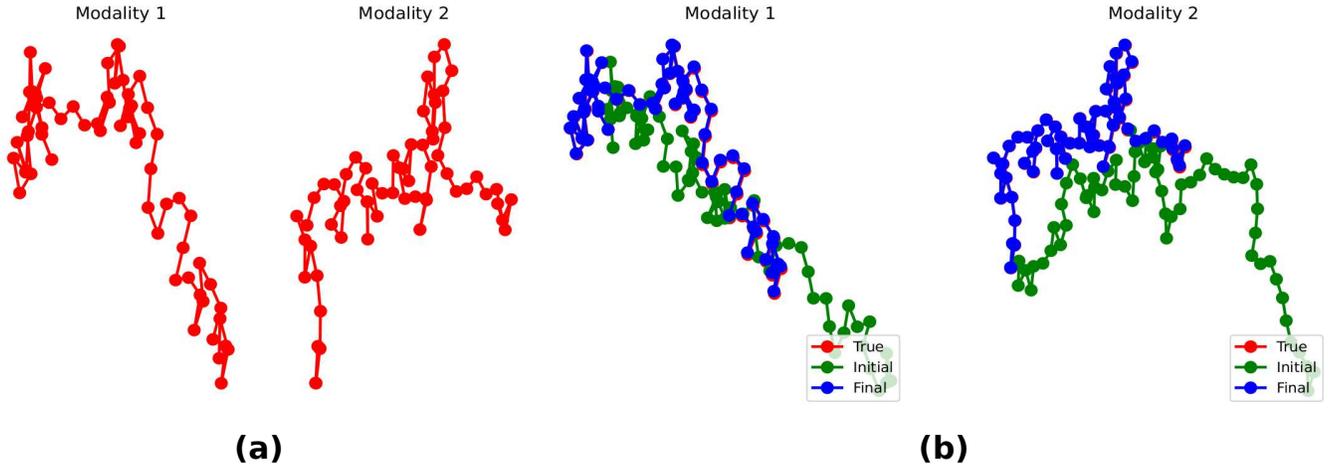} 
    \caption{(\textbf{a}) Generated toy proteins examples ($N=64$) and (\textbf{b}) reconstruction from compressed sensing observations. With 2 out of 64 measurements (3.125\%), near perfect reconstruction is possible even though the modalities are individually subsampled.}
    \label{fig:protein}
\end{figure}

A synthetic dataset containing two-dimensional (2D) protein-like data samples with two modalities is generated. The latent distribution $p(z), z\in\mathds{R}^4$ is a Gaussian mixture model with 10 components, simulating 10 different ``classes'' for samples. For each modality, the data generative model $p(x_m|z), x_m \in \mathds{R}^N$ is a one-layer multilayer perceptron (MLP) with random weights. Here $m=1,2$ represents two modalities. 10,000 pairs of samples are generated using the generative model, with the protein size $N=32$. 
Fig.~\ref{fig:protein}(a) shows an instance of the 2D protein data with $N=64$.

\subsection*{Passive WiFi radar dataset}
We use the OPERAnet \cite{operanet} dataset which was collected with the aim to evaluate human activity recognition (HAR) and localization techniques with measurements obtained from synchronized Radio-Frequency (RF) devices and vision-based sensors. The RF sensors captured the changes in the wireless signals while six daily activities were being performed by six participants, namely, sitting down on a chair ("sit"), standing from the chair ("stand"), laying down on the floor ("laydown"), standing from the floor ("standff"), upper body rotation ("bodyrotate), and walking ("walk").
We convert the raw time-series CSI data from the WiFi sensors into the image-like format, namely, spectrograms using signal processing techniques. 
More details are available in Section \ref{appendix_wifi_sp} of the Supplementary Information document.
2,906 spectrogram samples (each of 4s duration window) were generated for the 6 human activities and 80\% of these were used as training data while the remaining 20\% as testing data (random train-test split).

\section*{Results and Discussion}
\subsection*{Classification results of WiFi CSI spectrograms for HAR}
In this section, we evaluate the HAR sensor fusion classification performance under a few-shot learning scenario, with 1, 5 and 10 labelled examples per class. These correspond to 0.05\%, 0.26\% and 0.51\% of labelled training samples, respectively.
We randomly select 80\% of the samples in the dataset as the training set and the remaining 20\% is used for validation. The average $F_1$-macro scores for the HAR performance are shown in Table \ref{classify_results} for different 
models. To allow for a fair comparison, the same random seed was used in all experiments with only two modalities (processed spectrograms data obtained from two different receivers). 

Prior to training our model (see Supplementary Fig. \ref{fig:MVAEfullmodel}),
the spectrograms were reshaped to typical image dimensions of size $(1\times224\times224)$. Our model was trained for 1,000 epochs using the training data with a fixed KL scaling factor of $\beta = 0.02$. The encoders comprised of the ResNet-18 backbone with the last fully-connected layer dimension having a value of 512. For the decoders, corresponding CNN deconvolutional layers were used to reconstruct the spectrograms from each modality with the same input dimension. 
The latent dimension, batch size, and learning rate are set at 64, 64, and 0.001, respectively. 
In the second stage, the generative model serves as a reconstruction prior and the search manifold for the sensor fusion tasks. Essentially, in this stage, we obtain the maximum a posteriori estimate of $\hat{z}_{MAP}$ through the process described in Algorithm \ref{alg:cap}. The final estimate of the class is produced by $K$-NN in the latent representation space, with labelled examples sampled from the training set. 

To benchmark our technique we investigate the performance of other state-of-the-art sensor fusion techniques. The feature-fusion is represented by CNN models (1-channel CNN, 2-channel CNN, dual-branch CNN). All are trained in a conventional supervised fashion from scratch using the ResNet-18 backbone and a linear classification head is appended on top of it consisting of a hidden linear layer of 128 units and a linear output layer of 6 nodes (for classifying 6 human activities).
The dual-input CNN refers to the case where the embeddings from the two modalities’ CNNs are concatenated, and a classification head is then added (as illustrated in Fig. \ref{fig:SensorFusion}(b)). 
The “Probability Fusion” (decision fusion) model refers to a score-level fusion method where the classification probabilities ($P_1$ and $P_2$) from each modality are computed independently (using an output SoftMax layer) and then combined using the product rule (this is optimal given conditional independence).
These models 
are fine-tuned with labelled samples over 200 epochs, with a batch size of 64 and the Adam optimizer was used with learning rate of 0.0001, weight decay of 0.001 and $\beta_1= 0.95$, $\beta_2= 0.999$. 

It can be observed from Table \ref{classify_results} that our method significantly outperforms all other conventional feature and decision fusion methods. The confusion matrix for HAR classification using our SFLR (\underline{S}ensor \underline{F}usion in the \underline{L}atent \underline{R}epresentation
space) model is shown in Fig. \ref{cm_mvaenopoeknn} in the Supplementary Information document for the case when only ten labelled examples are used at the (classification) fusion stage.

\begin{figure}[t]
    \centering
    \includegraphics[width=15cm]{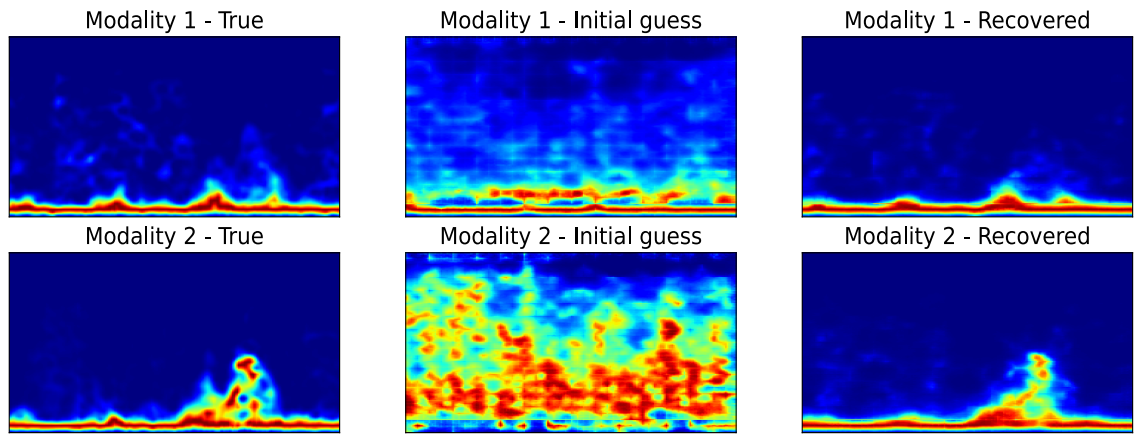}
    \caption{Illustration of spectrogram recovery (for  sitting down activity) using compressed sensing with measurements as low as 784  out of 50,176 (1.56\%). No additive white Gaussian noise is considered. The left column shows the true spectrogram sample, the middle column shows reconstruction with an initial guess (no optimization) while the right column shows reconstruction with $\hat{Z}_{MAP}$.}
    \label{cs_measurement_784}
\end{figure}

\subsection*{Sensor fusion from subsampled observations}
\label{sec:cs}
Next, we evaluate the recovery performance of spectrograms under different numbers of compressed sensing measurements. 
The measurement function $\chi_m$ is a matrix initialized randomly
and we assume that there is no additive Gaussian noise.
The Adam optimizer is used to optimize $\hat{z}_{MAP}$ with a learning rate of 0.01.
The algorithm is run for 10,000 iterations. 
After the loss in equation~\eqref{eq:map_loss} has converged during the optimization process, the samples are decoded/recovered for modality 1 and modality 2 using their respective decoders $\hat{x}_{m} = \psi_{m}(\hat{z}_{MAP})$.
Table \ref{tab_cs} shows the compressed sensing results when a batch of 50 images is taken from the testing dataset and evaluated under different number of measurements (without noise). It can be observed that the samples can be recovered with very low reconstruction error when the number of measurements is as low as 196 (0.39\%). 
An illustration is also shown in Fig. \ref{cs_measurement_784} where very good reconstruction is observed for the case when the number of measurements is equal to 784. More illustrations are shown in Fig. \ref{cs_exp_real_data} in the Supplementary Information document, with further experimental results in Sections \ref{noise_exp_sec}, \ref{mpr_exp_sec}, \ref{multimodal_exp_sec}.

\subsection*{Toy protein classification}
Similarly to the experiments on the OPERAnet dataset, we perform two tasks, classification and sensor fusion from compressed sensing observations, on the synthetic toy protein dataset.

As mentioned previously,
the toy protein dataset contains 10 classes. The dataset is split into a training set and a test set, containing 80\% and 20\% of samples, respectively. We evaluate the classification performance under a few-shot learning setting, using 1, 5 or 10 labelled samples per class.
The few-shot classification via the SFLR model consists of two stages. In the first stage, the M-VAE is trained in an unsupervised manner using the training set. Using the maximum a posterior $\hat{z}_{MAP}$ and a few labels, the $K$-NN classifier is applied to the latent representation space. Here the \textit{encoder} and \textit{decoder} in M-VAE are two-layer MLPs, with 16 neurons in the hidden layer. 

We compare the SFLR method with 4 baseline models. The single modality model only considers one modality without sensor fusion. The probability fusion model independently computes the classification probability for each modality, which is a representative model for decision-fusion (Fig.~\ref{fig:SensorFusion}(a)). The dual-branch feature fusion model concatenates the embedding of two modalities before the classification layer, which is a feature fusion method (Fig.~\ref{fig:SensorFusion}(b)). All baseline models are trained in a supervised manner, with the same neural network structure as the \textit{encoder}. Table~\ref{tab:protein_classification_results} shows the $F1$-macro scores for different methods on the test set. On the 10-class protein dataset, SFLR outperforms other sensor fusion models using limited labelled samples.

\subsection*{Sensor fusion from subsampled toy proteins}
\label{sec:cs_toy_protein}
Another advantage of the proposed SFLR model is that it can fuse modalities in subsampled cases. We use a set of \textit{samplers} $\chi_{1:M}$ to simulate the subsampled observations. The measurement function $\chi_m$ is a matrix initialized randomly. 
Here we use 10 initialization points to reduce the risk of getting trapped in a local minimum (points sampled from the prior with Stage 2 replicated for all of them).
Fig.~\ref{fig:protein}(b) shows the recovered protein from subsampled observations, with only 2 measurements for each modality. Both modalities are successfully recovered from the latent representation space, even though the initial guess $z^0$ is far from the true modality. Note that the proteins in Fig.~\ref{fig:protein} have a higher dimension than in the dataset, showing the robustness of the SFLR method. Table~\ref{tab:cs_recon_protein} shows the average reconstruction error of the synthetic protein dataset using different subsamplers. The reconstruction error reduced significantly when having 2 measurements for each modality, showing superior sensor fusion abilities.

The Supplementary Information document (see Section \ref{app:additional_res}) contains additional experiments, including tasks showcasing the ability to leverage between modalities, where a strong modality can be used to aid the recovery of a weak modality. It also presents the performance under subsampled and noisy conditions.

\section*{Conclusions and Broader Impacts} 
The paper presents a new method for sensor fusion. Specifically, we demonstrate the effectiveness of classification and reconstruction tasks from radar signals. The intended application area is human activity recognition, which serves a vital role in the E-Health paradigm. New healthcare technologies are the key ingredient to battling spiralling costs of provisioning health services that beset a vast majority of countries. Such technologies in a residential setting are seen as a key requirement in empowering patients and imbuing a greater responsibility for own health outcomes. However, we acknowledge that radar and sensor technologies also find applications in a military context. Modern warfare technologies (principally defensive) could potentially become more apt if they were to benefit from much-improved sensor fusion. We firmly believe that, on balance, it is of benefit to the society to continue the research in this area in the public eye.

\section*{Data availability}
The raw dataset used in the Passive WiFi radar experiments is available from \cite{operanet}. The toy protein dataset is not publicly available at this time but can be made available from the authors upon reasonable request.

\bibliography{sample}

\section*{Acknowledgements}
This work was performed as a part of the OPERA Project, funded by the UK Engineering and Physical Sciences Research Council (EPSRC), Grant EP/R018677/1. This work has also been funded in part by the Next-Generation Converged Digital Infrastructure (NG-CDI) Project, supported by BT and Engineering and Physical Sciences Research Council (EPSRC), Grant ref. EP/R004935/1.

\section*{Author contributions statement}
All authors, R.P, X.W and M.B, contributed equally to this work. The main tasks involved conceiving and conducting the experiments, algorithm implementation, analysis, validation and interpretation of results, and finally preparing and reviewing the manuscript. 

\section*{Additional information}
\subsection*{Competing interests}
The authors declare no competing interests.


\section*{Figure legends}
\begin{enumerate}
  \item Multimodal Sensor Fusion: (\textbf{a}) Decision fusion, (\textbf{b}) Feature fusion, (\textbf{c}) Our technique: fusion in the latent representation with optional compressed sensing measurements; $F$ features, $p(z)$ prior model, $\bf{G}$ generators, $X$ complete data, $Y$ subsampled data. For clarity $M=2$ modalities are shown, the concept generalises to any $M$.
  
  \item (\textbf{a}) Generated toy proteins examples ($N=64$) and (\textbf{b}) reconstruction from compressed sensing observations. With 2 out of 64 measurements (3.125\%), near perfect reconstruction is possible even though the modalities are individually subsampled.
  
  \item Illustration of spectrogram recovery (for  sitting down activity) using compressed sensing with measurements as low as 784  out of 50,176 (1.56\%). No additive white Gaussian noise is considered. The left column shows the true spectrogram sample, the middle column shows reconstruction with an initial guess (no optimization) while the right column shows reconstruction with $\hat{Z}_{MAP}$.
  
\end{enumerate}

\begin{table}[h]
\centering
\caption{\label{classify_results}  
Few-shot learning sensor fusion classification results ($F_1$ macro) for Human Activity Recognition.}
\begin{tabular}{c|c|c|c}
\hline
Model  & 1 example  & 5 examples & 10 examples  \\ \hline

2-channel CNN   & 0.427272 & 0.570888		 &	 0.618501                      \\ 
1-channel CNN (Modality 1)   & 0.349084 & 0.451328 &0.504462 
	                       \\ 
1-channel CNN (Modality 2)   & 0.446554 & 0.600084	& 0.605678	 
\\ 
Probability fusion (product rule)  & 0.440414 & 0.584726 & 0.641922 	\\         
Dual-branch CNN   & 0.508243        & 0.568795 &0.575914                      \\ 
SFLR (ours)    &  \textbf{0.652699}  & \textbf{0.718180} & \textbf{0.737507}  \\ 
\hline
\end{tabular}
\end{table}

\begin{table}[h]
\centering
\caption{\label{tab_cs}  
 Compressed sensing mean reconstruction error over a batch of 50 WiFi spectrogram data samples (No additive Gaussian noise). 
 An illustration is shown in Fig. \ref{cs_measurement_784}. 
 }
\begin{tabular}{l|l|l}
\hline
No. of measurements   & Modality 1    & Modality 2       \\ \hline
1 (0.002\%)    & 0.03118854         & 0.15024841                       \\ 
10 (0.02\%)   & 0.00938917         & 0.02824161                        \\ 
196 (0.39\%)  & 0.00348606         & 0.00613665                       \\ 
784 (1.56\%)  & 0.00305005         & 0.00505758                      \\ 
1,568 (3.125\%) & 0.00284343         & 0.00489433                      \\ \hline
\end{tabular}
\end{table}

\begin{table}[h]
\centering
\caption{\label{tab:protein_classification_results}  
Few-shot learning sensor fusion classification results ($F_1$ macro) for synthetic proteins.}
\begin{tabular}{c|c|c|c}
\hline
Model  & 1 example  & 5 examples & 10 examples  \\
\hline
Single modality (Modality 1) & 0.3188 & 0.4342 & 0.5843 \\

Single modality (Modality 2) & 0.3221 & 0.4849	& 0.5555 \\

Probability fusion (product rule)  & 0.2256 & 0.3736 & 0.3836 	\\

Dual-branch feature fusion & 0.3769 & 0.4973 & 0.5953        \\ 

SFLR (ours)    &  \textbf{0.4183}  & \textbf{0.5501} & \textbf{0.6120}  \\ 
\hline
\end{tabular}
\end{table}

\begin{table}[h]
\caption{Compressed sensing mean reconstruction error over a batch of 100 protein samples.}
    \centering
    \begin{tabular}{c|c|c}
    \hline
     No. of Measurements & Modality 1 ($10^{-5}$) & Modality 2 ($10^{-5}$)\\
\hline
     1 (3.125\%)  & 4,622.4 & 4,923.5 \\
     2 (6.250\%) & 22.5 & 27.9 \\
     4 (12.500\%) & 7.1 & 7.4 \\
     8 (25.000\%) & 2.3 & 2.7 \\
\hline
    \end{tabular}
    \label{tab:cs_recon_protein}
\end{table}


\beginsupplement
\maketitle

\clearpage
\section*{Appendix}
\vspace{0.5em}

\section{Approximations to variational posterior}
\label{App:PoE}
Given the objective, the variational joint posterior $q_\phi (z|x_{1:M})$ can be learned by training one single encoder network that takes all modalities $X_{1:M}$ as input to explicitly parametrize the joint posterior. This is our baseline model and an example for $M=2$ modalities is given in Fig.~\ref{fig:MVAEfullmodel}. However, this approach requires all modalities to be present at all times, thus making cross-modal generation difficult. 
Alternatively, the joint variational posterior can be modelled using the following approaches:
 
\subsection*{Variational Product of Experts}
In this section we reproduce the arguments from \cite{mvae}. The first option is to approximate the joint variational posterior as a product: 
\begin{equation}
q_{\phi}(z|x_{1:M}) \equiv p(z) \prod_{m=1}^M{q_{\phi_m}(z|x_{m})}   .  
\end{equation}
In case of a missing expert, we assume: $q_{\phi_m}(z|x_m) = 1$.


For a system of $N$ modalities, $2^N$ inference networks need to be specified, $q(z|X)$ for each subset of modalities $X\subseteq \{X_1,X_2,\dots, X_M \}$.
The optimal inference network $q(z|x_1,\dots, x_N)$ would be the
true posterior $p(z|x_1,\dots, x_N)$. The conditional independence assumptions in the generative model imply a relation among joint- and single-modality posteriors \cite{mvae}: 
\begin{align}
p(z|x_1,\dots, x_N) &=  \frac{p(x_1,\dots, x_N |z) p(z) }{p(x_1,\dots, x_N )} \nonumber \\ 
&=  \frac{p(z)}{p(x_1,\dots, x_N )} \prod_{i=1}^N p(x_i|z) \nonumber \\
&=  \frac{p(z)}{p(x_1,\dots, x_N )} \prod_{i=1}^N \frac{p(z|x_i) p(x_i)}{p(z)} \nonumber \\
&=  \frac{\prod_{i=1}^N  p(z|x_i)}{\prod_{i=1}^{N-1} p(z)} \cdot \frac{\prod_{i=1}^N p(x_i)}{p(x_1,\dots, x_N )} \nonumber\\
&\propto \frac{\prod_{i=1}^N  p(z|x_i)}{\prod_{i=1}^{N-1} p(z)}.
\label{joint_post_eq}
\end{align}

If we approximate $p(z|x_i)$ with $q(z|x_i)\equiv \tilde{q}(z|x_i)p(z)$, where $\tilde{q}(z|x_i)$ is the underlying inference network, the quotient term can be omitted \cite{mvae}: 
\begin{align}
p(z|x_1,\dots, x_N) & \propto \frac{\prod_{i=1}^N  p(z|x_i)}{\prod_{i=1}^{N-1} p(z)} \nonumber \\
&\approx \frac{\prod_{i=1}^N [\tilde{q}(z|x_i)p(z)]}{\prod_{i=1}^{N-1} p(z)} \nonumber \\
&= p(z) \prod_{i=1}^N \tilde{q}(z|x_i).
\label{joint_post_eq2}
\end{align}
Equation (\ref{joint_post_eq2}) implies that we can use a Product-of-Experts (PoE), including a “prior expert” (e.g., spherical Gaussian), as the approximating distribution for the joint-posterior. This derivation is easily extended to any subset of modalities yielding $q(z|X) \propto p(z) \prod_{x_i \in X}\tilde{q}(z|x_i)$.

\subsection*{Variational Mixture of Experts}
\begin{equation}
q_{\phi}(z|x_{1:M}) \equiv \sum_{m=1}^M{\frac{1}{M} q_{\phi_m}(z|x_{m})}      ,
\end{equation} 
where 
the above assumes an equitable distribution of power among experts. Non-uniform weights can also be used.   
Missing expert: $q_{\phi_m}(z|x_m) = 0$.

\begin{figure}[t]
    \centering
    \includegraphics[width=15cm]{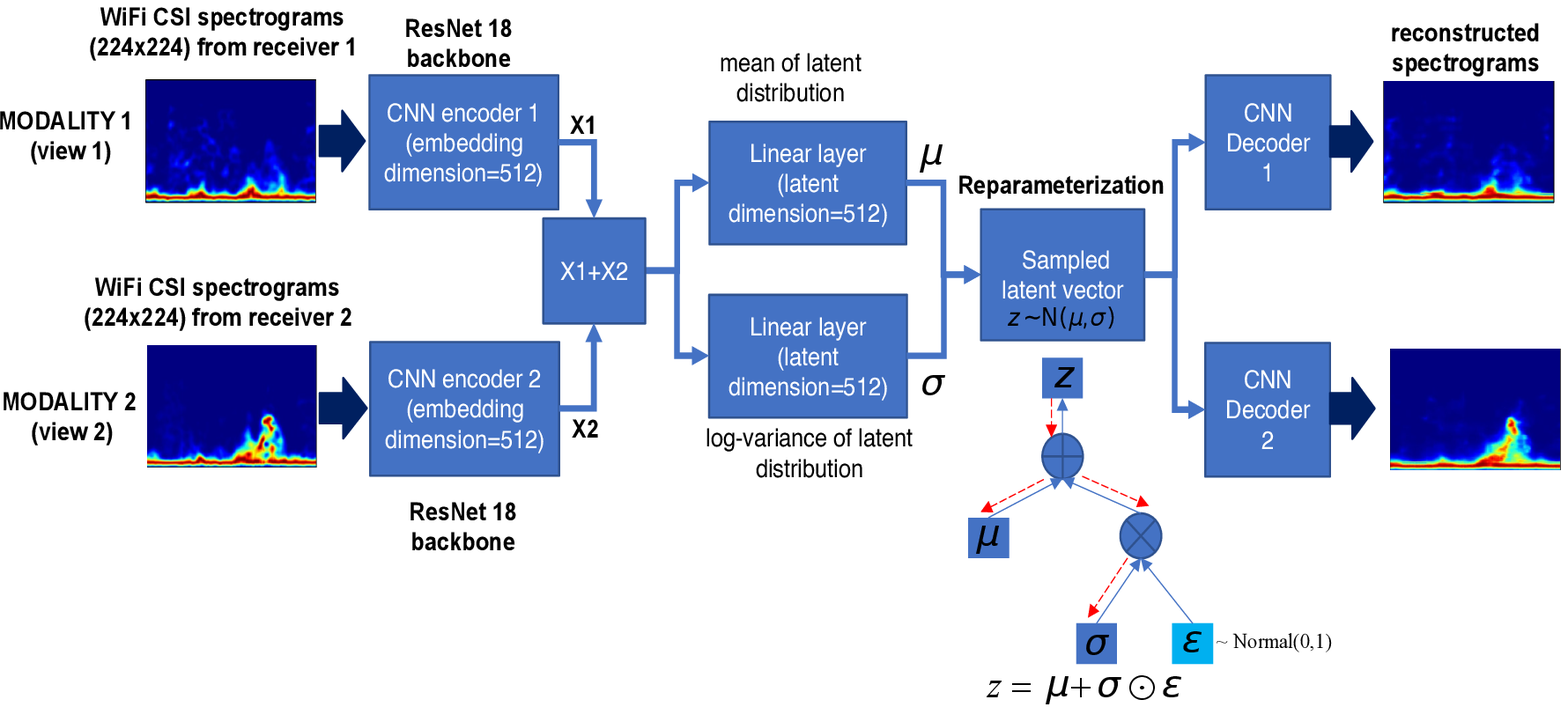}
    \caption{M-VAE for a full data case: Single encoder network takes all modalities.}
    \label{fig:MVAEfullmodel}
\end{figure}

\section{Signal processing pipelines for passive WiFi radar}
\label{appendix_wifi_sp}
\begin{figure}[t]
    \centering
    \includegraphics[width=11cm]{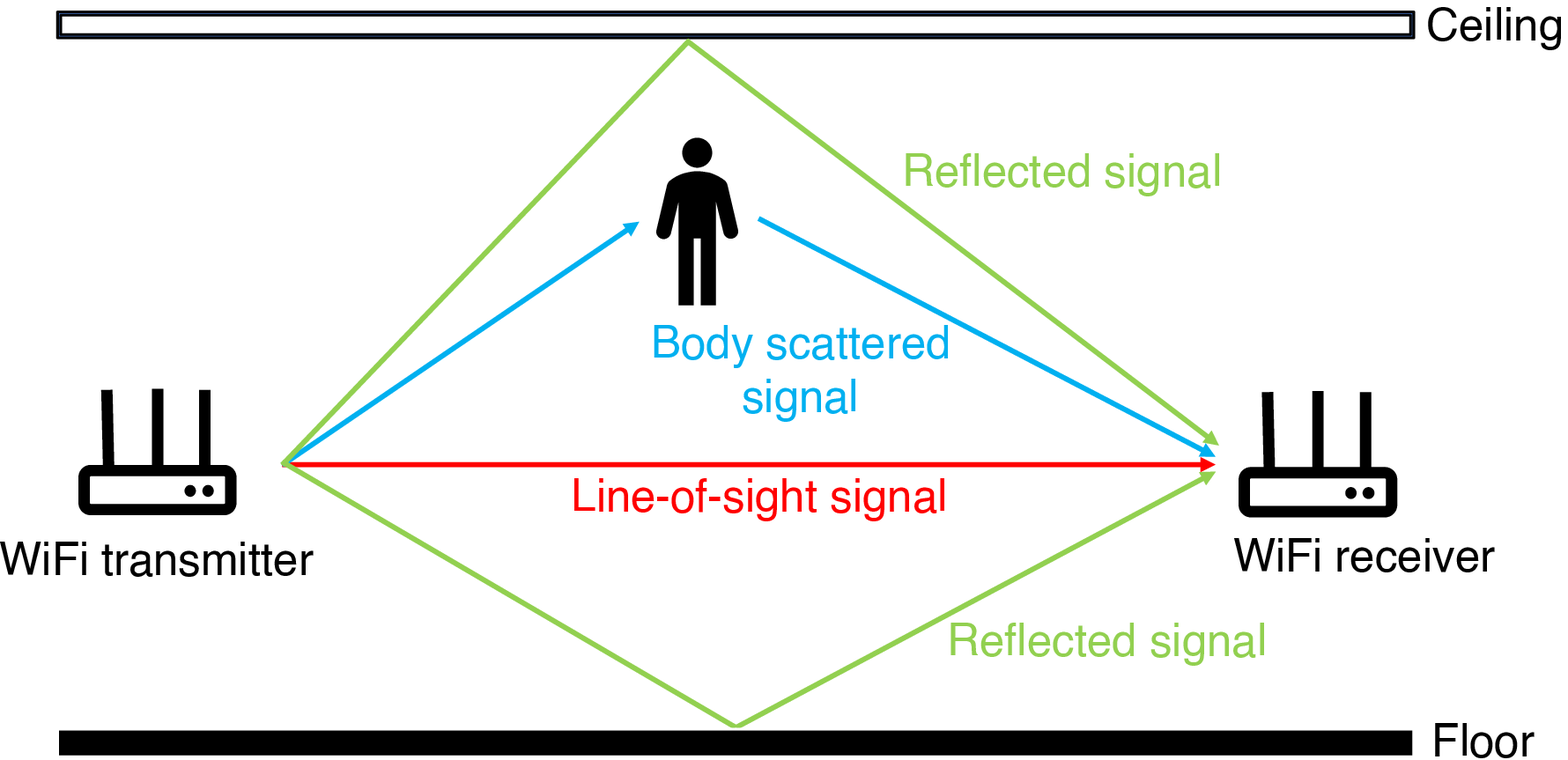}
    \caption{Opportunistic Passive WiFi Radar.}  
    \label{fig:WiFiImage}
\end{figure}

A typical scenario for opportunistic passive WiFi Radar is depicted in Figure \ref{fig:WiFiImage}. This is an extremely challenging problem since the WiFi waveform was not specifically designed to lend itself to Radio-Frequency (RF) imaging. In addition, commercial WiFi chipsets have noisy RF chains and tend to suffer from phase drifts. The WiFi backscatter does contain information about the dynamic changes in the radio channel which is incapsulated in the Channel State Information (CSI).  
Dedicated tools need to be used to extract the CSI from WiFi network interface cards such as Atheros \cite{atheros_ref} or Intel 5300 (IWL5300) \cite{csi_tool}. The raw CSI data is obtained as a 3-dimensional (3D) matrix per transmitted packet, with $n_t$$\times$$n_r$$\times$$N_{\text{sc}}$ complex values, where $n_t$ is the number of transmit antennas, $n_r$ is the number of receive antennas and $N_{\text{sc}}$ is the number of subcarriers.
Since the raw CSI data is very noisy in nature, the Discrete Wavelet Transform (DWT) technique can be used to filter out in-band noise and preserve the high frequency components, thus avoiding the distortion of the signal \cite{translation_resilient}. Afterwards, median filtering can be used to remove any undesired transients in the CSI measurements which are not due to human motion. 
The Intel 5300 chipset has a 3$\times$3 antenna configuration and only 30 subcarriers are reported by this chipset. Thus the number of complex values per packet is equal to $3$$\times$$3$$\times$$30=270$. Considering a packet rate as high as 1.6 kHz, this results in a significant amount of data that needs to be processed. 
Therefore, we also apply Principal Component Analysis (PCA) to reduce the computational complexity of such high dimensional data. PCA identifies the time-varying correlations between the CSI streams which are optimally combined to extract only a few components that represent the variations caused by human activities.
Finally, we convert the resultant data into spectrograms (time-frequency domain) using Short Time Fourier Transform (STFT), which are similar to those generated by Doppler radars.
The CSI is highly sensitive to the surrounding environment
and signal reflections from the human body result in different
frequencies when performing different activities. The Doppler spectrogram generated from STFT helps to identify the change of frequencies over time. The generated spectrograms can be directly fed to CNNs to automatically identify a set of features, which can ultimately be used in downstream tasks.
The CSI system consisted of two receivers. For more details on the experimental setup of the data collection, the interested reader is kindly referred to \cite{operanet}. Each receiver can be seen as another view of the human activity being performed in the environment. 

\section{Latent representation of WiFi spectrogram data}
\begin{figure}[t]
    \centering
    \includegraphics[width=15cm]{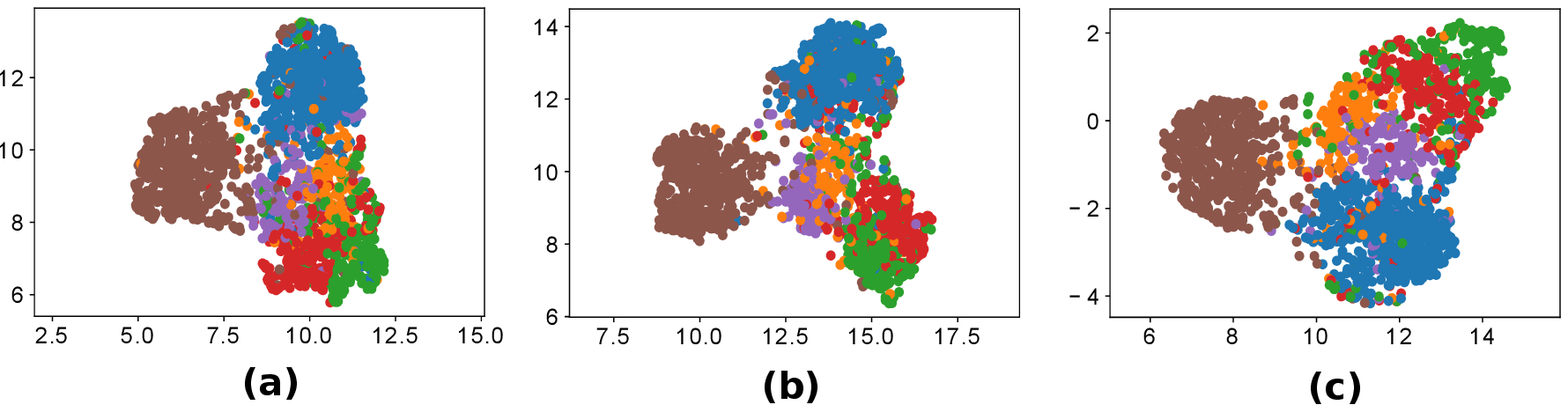}
    \caption{UMAP projection of trained latent space using our model on real WiFi CSI spectrogram data: (\textbf{a}) latent dimension=16, (\textbf{b}) latent dimension=64, and (\textbf{c}) latent dimension=128.}  
    \label{trainedlatent_realdata}
\end{figure}

Our base model is shown in Fig.~\ref{fig:MVAEfullmodel}.
The trained latent space for different latent dimensions
are shown in Fig.~\ref{trainedlatent_realdata}. 
The trained latent space in Fig.~\ref{trainedlatent_realdata} shows distinct clusters using UMAP (Uniform Manifold Approximation and Projection) visualization. The model was trained in a self-supervised fashion, the six clusters representing the six different human activities can be seen.

\section{Sensor fusion under noisy conditions (WiFi spectrogram data)}
\label{noise_exp_sec}
\begin{figure}[t]
    \centering
    \includegraphics[width=12cm]{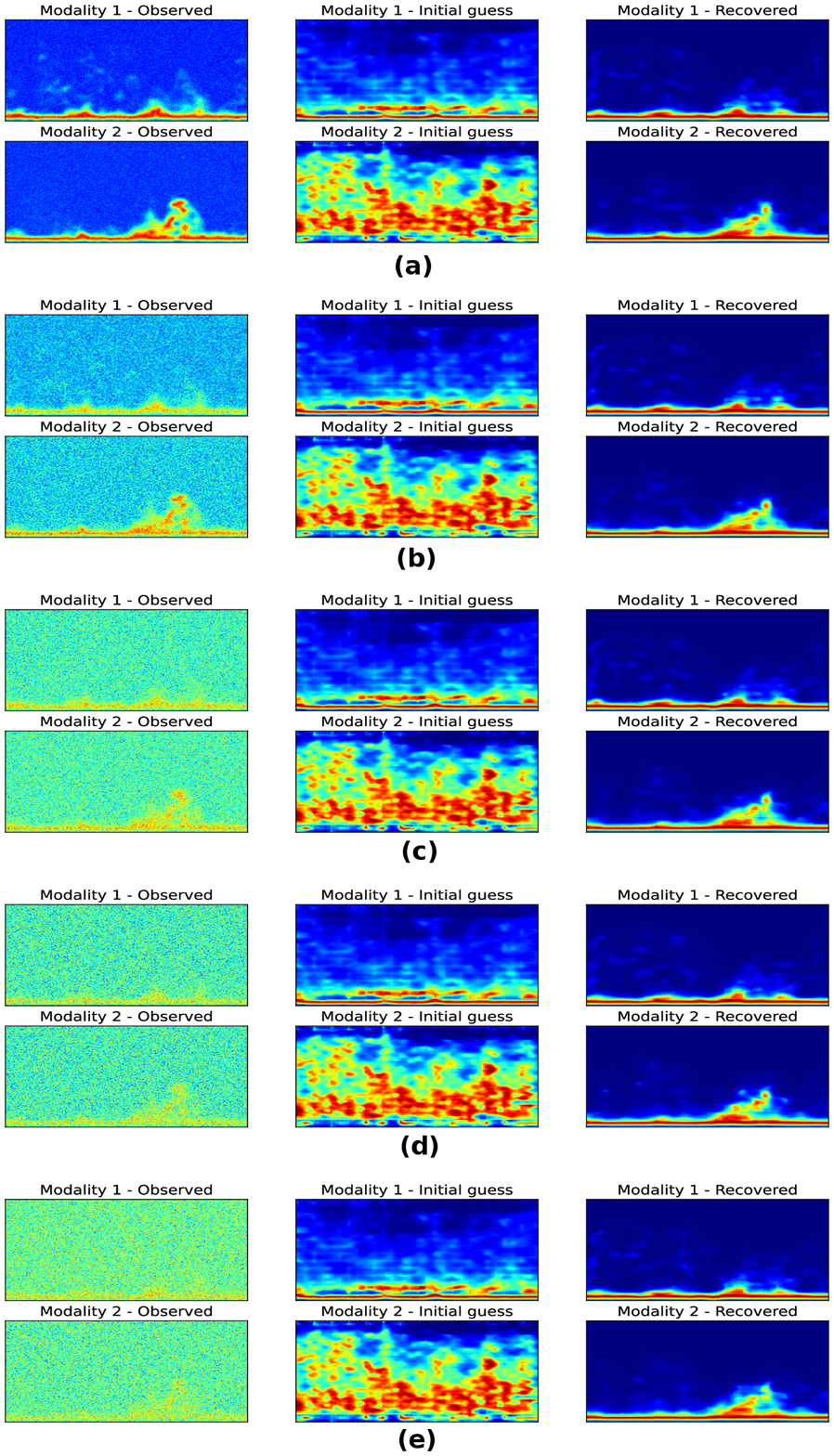}
    \caption{Impact of additive Gaussian noise on sensor fusion from the two modalities: (\textbf{a}) Std Dev=0.05, (\textbf{b}) Std Dev=0.2, (\textbf{c}) Std Dev=0.4, (\textbf{d}) Std Dev=0.6, and (\textbf{e}) Std Dev=0.8. Left column shows noisy spectrogram sample, middle column shows fusion with initial guess (no optimization) while right column shows fusion with $\hat{z}_{MAP}$.}  
    \label{exp_denoise}
\end{figure}

In this experiment, we analyze the sensor fusion performance when the data samples from the test dataset are affected by different amount of additive Gaussian noise.
In this case, the measurement matrices 
are initialized as identity matrices with dimensions 50,176$\times$50,176. 
No noise was injected to the input data from two modalities during training.
The SFLR (\underline{S}ensor \underline{F}usion in the \underline{L}atent \underline{R}epresentation
space) algorithm (see Algorithm 1 in manuscript) is run for 1,000 iterations and the corresponding results are shown in Fig.~\ref{exp_denoise}, showing one sample in the test dataset. It can be observed that even under extreme noisy conditions, the noisy samples are denoised efficiently. These results are further validated in Table \ref{noise_tab} where it can be seen that the fusion error remains essentially constant for different noise standard deviation values considering a batch of 50 images from the test dataset.

\begin{figure}[t]
    \centering
    \includegraphics[width=12cm]{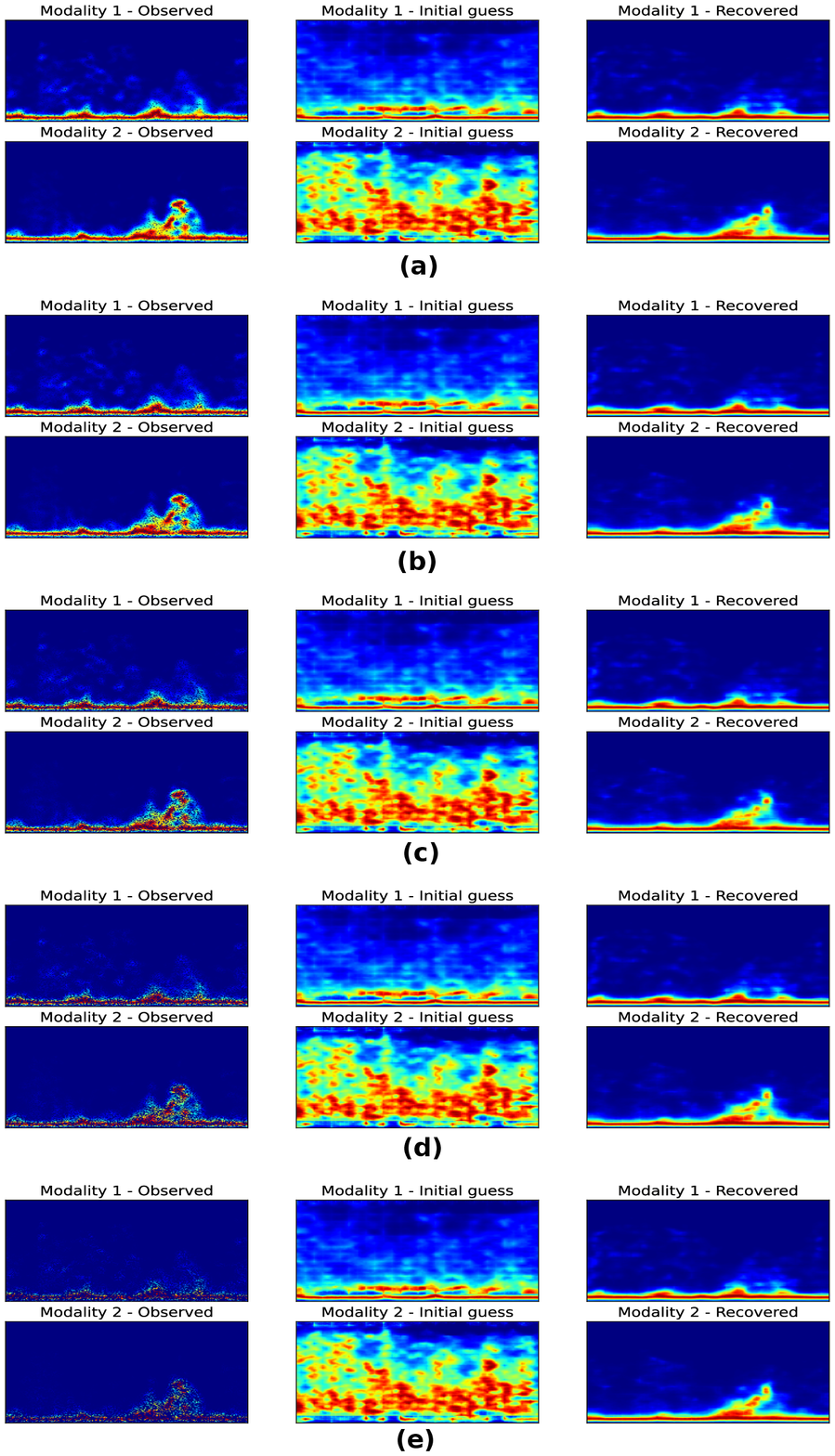}
    \caption{Impact of missing pixels on spectrogram recovery from the two modalities (no additive Gaussian noise): (\textbf{a}) missing pixel ratio=0.1, (\textbf{b}) missing pixel ratio=0.2, (\textbf{c}) missing pixel ratio=0.4, (\textbf{d}) missing pixel ratio=0.6, and (\textbf{e}) missing pixel ratio=0.8.   
    Left column shows spectrogram sample with missing pixels, middle column shows reconstruction with initial guess (no optimization) while right column shows reconstruction with $\hat{z}_{MAP}$. Very good recovery performance is observed in all cases.}  
    \label{exp_mpr}
\end{figure}

\begin{table}[h]
\centering
\caption{\label{noise_tab} 
Noisy measurements mean reconstruction error over a batch of 50 WiFi spectrogram data samples (full measurements considered).} 

\begin{tabular}{l|l|l}
\hline
Noise standard deviation  & Modality 1    & Modality 2       \\ \hline
0.01    & 0.00347822          & 0.00573636                     \\ 
0.05   & 0.00348499         & 0.00574255                       \\ 
0.20   & 0.00349233         & 0.00569815                       \\ 
0.40   & 0.00354153         & 0.0058348                        \\ 
0.60   & 0.00357407        & 0.00594901                      \\ 
0.80   & 0.00367526        & 0.00608692                        \\ \hline
\end{tabular}
\end{table}


\section{Sensor fusion performance with missing pixels (WiFi spectrogram data)}
\label{mpr_exp_sec}
In this experiment, we evaluate the fusion performance of the samples under different ratios of missing pixels. The results are shown in Fig.~\ref{exp_mpr} when a true data sample is randomly chosen from the test dataset and a randomly generated (binary) mask is applied to it to simulate different missing pixel ratios. 
Each measurement corresponds to an observed pixel.
Therefore, in this case the measurement matrices 
will be diagonal matrices with their diagonal entries corresponding to the mask elements (1's and 0's). From Fig.~\ref{exp_mpr}, it can be observed that the recovered samples are very close to the true ones, even when the missing pixel ratio for both modalities is as high as 0.8. In fact, as shown in Table \ref{tab_mpr}, the reconstruction error remains essentially constant with increasing missing pixel ratio for both modalities. 

\begin{table}[ht]
\centering
\caption{\label{tab_mpr} 
 Missing pixel mean reconstruction error over a batch of 50 WiFi spectrogram data samples. Illustrations of spectrogram fusion under different missing pixel ratios are shown in Fig. \ref{exp_mpr}.} 
\begin{tabular}{l|l|l}
\hline
Missing pixel ratio   & Modality 1    & Modality 2       \\ \hline
0.1   & 0.00351316          & 0.00573596                       \\ 
0.2   & 0.00345235         & 0.00575154                      \\ 
0.4   & 0.00347854         & 0.00575382                       \\ 
0.6   & 0.003477        & 0.00571794                        \\ 
0.8   & 0.00349462         & 0.00579462                      \\ 
\hline
\end{tabular}
\end{table}

\section{Sensor fusion from asymmetric Compressed Sensing (WiFi spectrogram data)}
\label{multimodal_exp_sec}
In this experiment, we analyse the reconstruction of the WiFi spectrogram samples under two different scenarios, where we want to demonstrate the benefits of having multiple modalities. We are interested in recovery for one modality that is subsampled (loss of data) and noisy. This can be referred to as the weak data (or weak modality). Using the SFLR method, we leverage the second modality data, which has no loss of information or does not suffer from noise (strong modality), to improve the recovery for the modality of interest i.e., the weak modality. 
In the first case, only modality 1 (subsampled and noisy) is considered in the reconstruction process. 
In the second case, the good modality 2 is added in the iterative fusion process to improve the reconstruction quality of modality 1.

\begin{figure}[t]
    \centering
    \includegraphics[width=12cm]{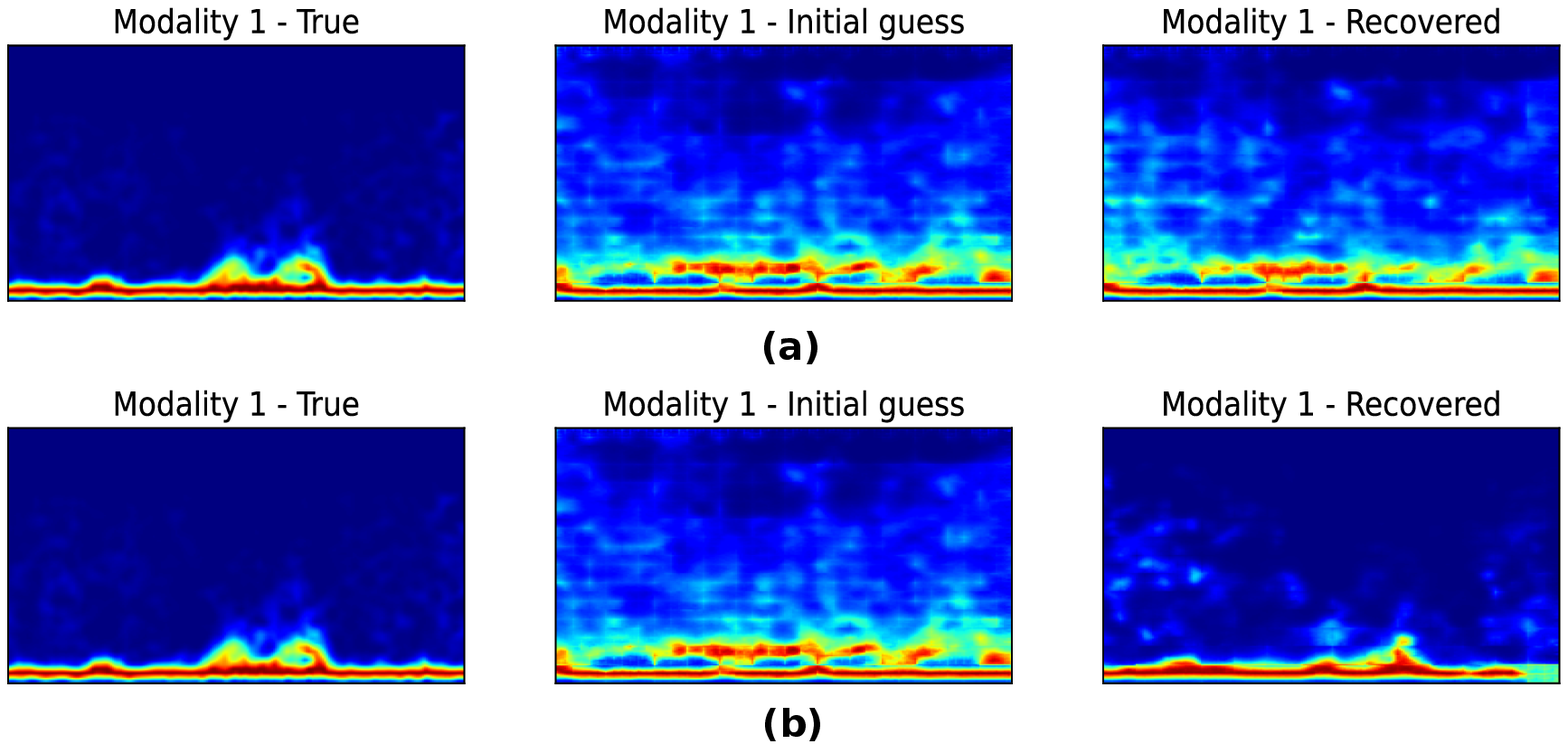}
    \caption{Reconstruction examples showing the benefit of multimodal system compared to a unimodal system. Modality 1 is subsampled data with 1 single measurement while modality 2 has full information (no noise and no loss of data). Additive Gaussian noise with a standard deviation of 0.1 is considered in this example: (\textbf{a}) reconstruction with modality 1 only, (\textbf{b}) reconstruction with both modalities 1 and 2.    
    Left column shows true spectrogram sample, middle column shows reconstruction with initial guess (no optimization) while right column shows reconstruction with $\hat{z}_{MAP}$. Adding modality 2 during reconstruction stage helps in the sample recovery of modality 1.}  
    \label{fig:claim3}
\end{figure}

The results are tabulated in Table \ref{label_claim3_wifi}, where additive Gaussian noise with a standard deviation of 0.1 is considered. The results show the mean reconstruction errors (over 50 WiFi spectrogram samples) when modality 1 is subsampled to different extents. 
We see that reconstruction error has a general tendency to decrease with increasing number of measurements. It can be observed that the samples can be recovered with very low reconstruction error when the number of measurements is as low as 196 (0.39\%). Furthermore, from Table~\ref{label_claim3_wifi}, we observe that when only modality 1 is considered in the reconstruction process, the reconstruction errors  are high when the number of measurements is equal to 1 (0.002\%) and 10 (0.02\%). However, by leveraging the good modality 2, the reconstruction quality is greatly improved for the same number of measurements, demonstrating the clear benefit of having multiple modalities. An illustration of the reconstruction quality is depicted in Fig. \ref{fig:claim3}, where it can be observed that the unimodal reconstruction of modality 1 is far from the true sample. On the other hand, the reconstruction quality of modality 1 is improved by leveraging the good modality data. 

\begin{table}[htb]
    \centering
        \caption{ Mean reconstruction error over 50 WiFi spectrogram data samples. Noise standard deviation: $0.1$} 
    \begin{tabular}{c|c|c|c}
    \hline
    & No. of Measurements & Modality 1 & Modality 2\\
\hline
    \multirow{5}{*}{Modality 1 with compressed sensing} & 
    1 (0.002\%)        &  0.0246185  & - \\
    & 10 (0.02\%)    &  0.01075371 & - \\
    & 196 (0.39\%)   &  0.00258467 & - \\
    & 784 (1.56\%)   &  0.00195997 & - \\
    & 1,568 (3.125\%) &  0.00184247 & - \\
\hline
    \multirow{3}{*}{Modality 1 with compressed sensing} &
    1 (0.002\%)  &  0.00892453 & 0.00380795 \\
    & 10 (0.02\%) & 0.00798366 & 0.00420512 \\
    & 196 (0.39\%) &   0.0034269 & 0.00460956 \\
    \multirow{2}{*}{Modality 2 with full information} & 
    784 (1.56\%) &  0.0030373 & 0.00466936 \\
    & 1,568 (3.125\%) &  0.0028537 & 0.00469946 \\
\hline
    \end{tabular}
    \label{label_claim3_wifi}
\end{table}

\begin{figure}[t]
    \centering
    \includegraphics[width=12cm]{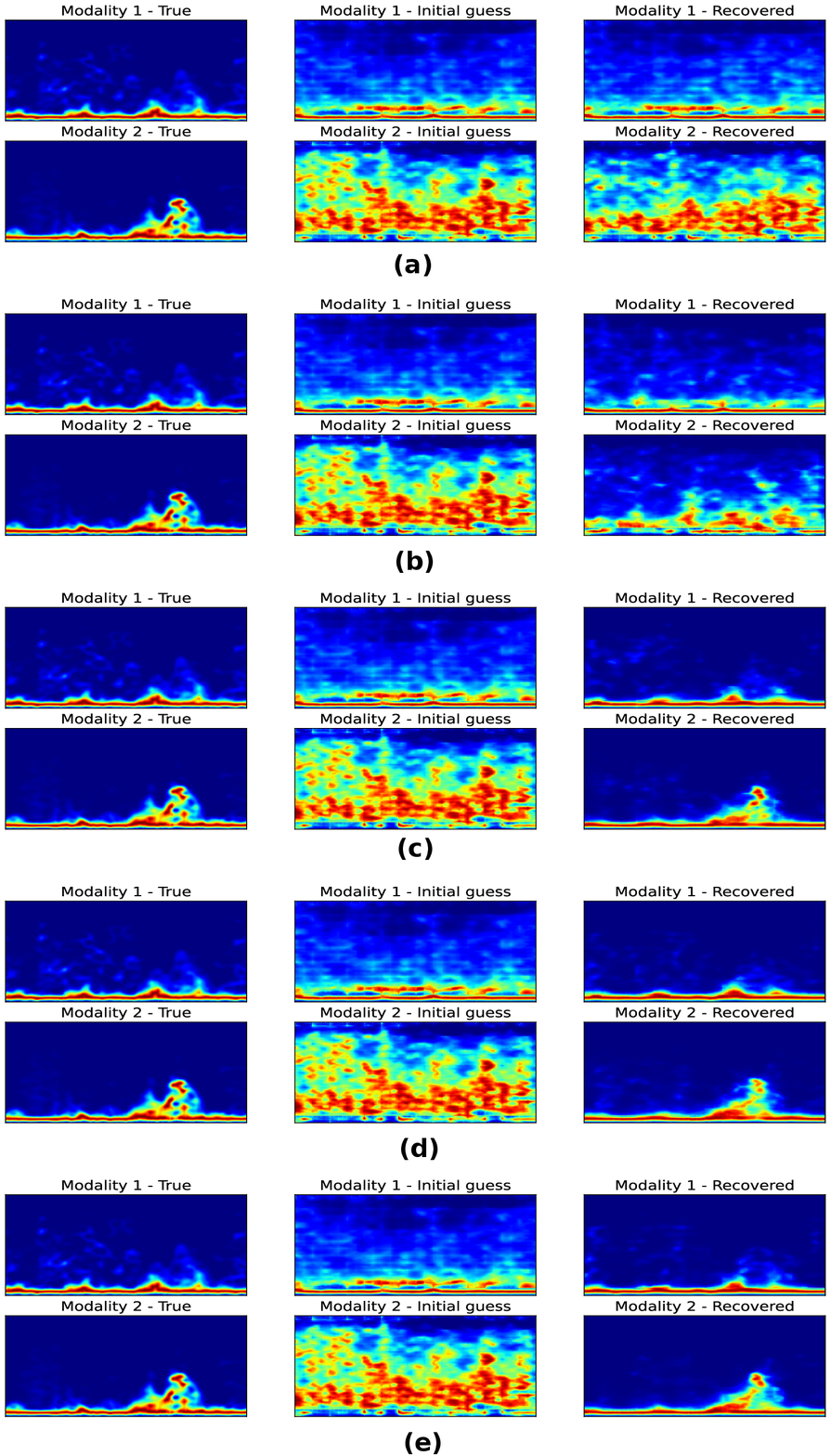}
    \caption{Compressed sensing performance on different number of measurements without additive Gaussian noise: (\textbf{a}) 1 measurement out of 50,176 (0.002\%), (\textbf{b}) 10 measurements out of 50,176 (0.02\%), (\textbf{c}) 196 measurements out of 50,176 (0.39\%), (\textbf{d}) 784 measurements out of 50,176 (1.56\%), and (\textbf{e}) 1,568 measurements out of 50,176 (3.125\%).     
 Left column shows true spectrogram sample, middle column shows reconstruction with initial guess (no optimization) while right column shows reconstruction with $\hat{z}_{MAP}$.}  
    \label{cs_exp_real_data}
\end{figure}

\section{Toy protein dataset: additional results }\label{app:additional_res}

\subsection*{Sensor fusion from subsampled and noisy toy proteins}
In this section, we present the sensor fusion results for toy protein reconstruction under subsampled and noisy observations, as an extension to Section "Sensor fusion from subsampled toy proteins" in the main document. Table~\ref{tab:cs_noisy_protein} shows the mean reconstruction error of subsampled toy protein samples, with different levels of additive Gaussian noise. The proposed SFLR method recovers both modalities from as low as 4 subsampled and noisy observations.

\begin{table}[htb]
    \caption{Compressed sensing mean reconstruction error over a batch of 100 protein samples, with different noise levels.}
    \label{tab:cs_noisy_protein}
    \centering
    \begin{tabular}{c|c|c|c}
    \hline
    Noise standard deviation & No. of Measurements &  Modality 1 ($10^{-3}$) & Modality 2 ($10^{-3}$) \\
    \hline
    \multirow{5}{*}{0.05} & 1 (3.125\%)  & 52.301 & 55.382 \\
    & 2 (6.250\%)  & 11.678 & 9.200 \\
    & 4 (12.500\%) & 0.834 & 0.715 \\
    & 8 (25.000\%) & 0.387 & 0.450 \\
    \hline
    \multirow{5}{*}{0.1} & 1 (3.125\%)  & 36.611 & 50.118 \\
    & 2 (6.250\%)  & 20.267 & 14.638 \\
    & 4 (12.500\%) & 3.413 & 2.769 \\
    & 8 (25.000\%) & 2.386 & 2.411 \\
    \hline
    \multirow{5}{*}{0.2} & 1 (3.125\%)  & 43.466 & 48.271 \\
    & 2 (6.250\%)  & 17.864 & 19.435 \\
    & 4 (12.500\%) & 1.528 & 1.435 \\
    & 8 (25.000\%) & 5.005 & 5.063 \\
    \hline
    \end{tabular}
\end{table}

\subsection*{Sensor fusion from asymmetric Compressed Sensing of toy proteins}
We show the results of sensor fusion from asymmetric compressed sensing, regarding the third contribution of this paper. We claim that a strong modality can be used to aid the recovery of another modality that is lossy or less informative (weak modality). Table~\ref{tab:asymmetric cs protein} shows the recovery results in two cases. In the first case, the subsampled modality 1 with additive Gaussian noise is observed and recovered. In the second case, the noise-free modality 2 with full observation is used to help the sensor fusion. We can see that modality 2 significantly helps with the recovery of modality 1, especially when the number of observations are relatively small.

\begin{table}[htb]
    \centering
    \caption{Mean reconstruction error over 100 samples with asymmetric compressed sensing. Noise standard deviation: $0.1$.}
    \begin{tabular}{c|c|c|c}
    \hline
    & No. of Measurements & Modality 1 & Modality 2\\
\hline
    \multirow{4}{*}{Modality 1 with compressed sensing} & 1 (3.125\%)  & 0.0542 & - \\
    & 2 (6.250\%) &  0.0366 & - \\
    & 4 (12.500\%) &  0.0205 & - \\
    & 8 (25.000\%) &  0.0021 & - \\
\hline
    \multirow{2}{*}{Modality 1 with compressed sensing} & 1 (3.125\%)  & 0.0076  & 0.0073 \\
    & 2 (6.250\%) & 0.0067 & 0.0062 \\
    \multirow{2}{*}{Modality 2 with full information} & 4 (12.500\%) & 0.0023 & 0.0024\\
     & 8 (25.000\%) & 0.0033  & 0.0031\\
\hline
    \end{tabular}
    \label{tab:asymmetric cs protein}
\end{table}


 \label{FirstAppendix}

\begin{figure}[t]
    \centering
    \includegraphics[width=10cm]{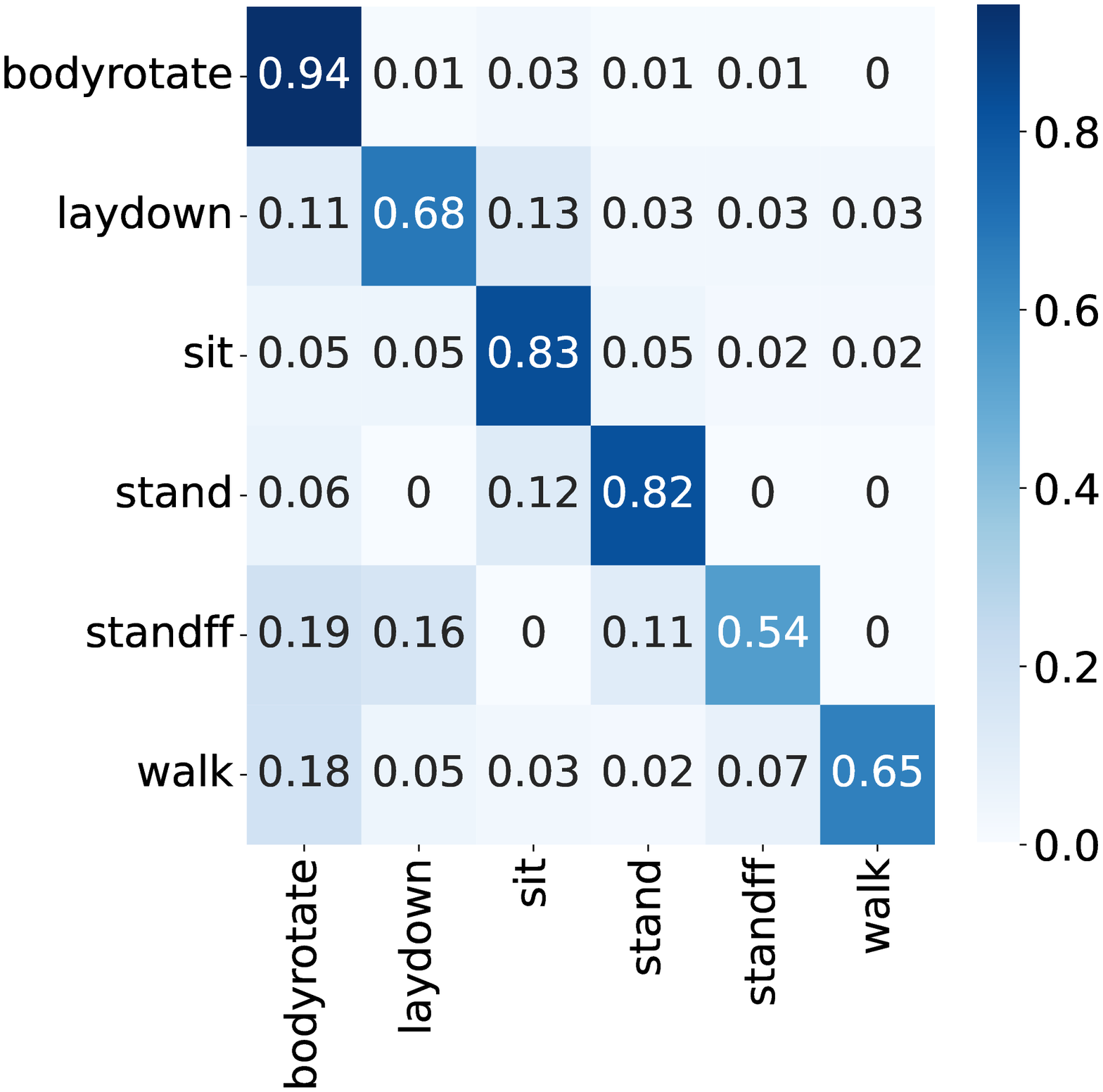}
    \caption{Confusion matrix of Human Activity Recognition (HAR) classification using our SFLR model (with compressed sensing). Ten labelled examples per class are considered (refer to Table 1 in main manuscript for classification results in terms of macro $F_1$ score).  
}
    \label{cm_mvaenopoeknn}
\end{figure}

\end{document}